\documentclass[10pt,twocolumn,letterpaper]{article}

\usepackage[pagenumbers]{cvpr} %

\usepackage{graphicx}
\usepackage{amsmath}
\usepackage{amssymb}
\usepackage{booktabs}
\usepackage[table]{xcolor}
\usepackage{pifont}%
\usepackage{array}
\usepackage{enumitem}
\usepackage{makecell}
\usepackage{amssymb}
\usepackage{mathtools}
\usepackage{amsthm}
\usepackage{multirow}

\usepackage[pagebackref,breaklinks,colorlinks]{hyperref}

\newcommand{\cmark}{{\color{black}\ding{51}}}
\newcommand{\xmark}{{\color{gray}\ding{55}}}
\newcommand{\tfbestfirst}[1]{\textcolor{magenta}{\textbf{#1}}}

\usepackage[capitalize]{cleveref}
\crefname{section}{Sec.}{Secs.}
\Crefname{section}{Section}{Sections}
\Crefname{table}{Table}{Tables}
\crefname{table}{Tab.}{Tabs.}

\def\cvprPaperID{5613} %
\def\confName{CVPR}
\def\confYear{2023}

\begin{document}

\title{Visual Exemplar Driven Task-Prompting for Unified Perception in Autonomous Driving}

\author{Xiwen Liang$^1$, Minzhe Niu$^2$, Jianhua Han$^2$, Hang Xu$^2$, Chunjing Xu$^2$, Xiaodan Liang$^1$$^\dagger$\\
$^1$Shenzhen Campus of Sun Yat-sen University, $^2$Huawei Noah's Ark Lab \\
{\tt\small\{liangxw29@mail2, liangxd9@mail\}.sysu.edu.cn,} \\
{\tt\small\{niuminzhe1, hanjianhua4, xu.hang, xuchunjing\}@huawei.com}
}
\maketitle

\begin{abstract}
Multi-task learning has emerged as a powerful paradigm to solve a range of tasks simultaneously with good efficiency in both computation resources and inference time. However, these algorithms are designed for different tasks mostly not within the scope of autonomous driving, thus making it hard to compare multi-task methods in autonomous driving. Aiming to enable the comprehensive evaluation of present multi-task learning methods in autonomous driving, we extensively investigate the performance of popular multi-task methods on the large-scale driving dataset, which covers four common perception tasks, i.e., object detection, semantic segmentation, drivable area segmentation, and lane detection. We provide an in-depth analysis of current multi-task learning methods under different common settings and find out that the existing methods make progress but there is still a large performance gap compared with single-task baselines.
To alleviate this dilemma in autonomous driving, we present an effective multi-task framework, VE-Prompt, which introduces visual exemplars via task-specific prompting to guide the model toward learning high-quality task-specific representations. Specifically, we generate visual exemplars based on bounding boxes and color-based markers, which provide accurate visual appearances of target categories and further mitigate the performance gap.
Furthermore, we bridge transformer-based encoders and convolutional layers for efficient and accurate unified perception in autonomous driving.
Comprehensive experimental results on the diverse self-driving dataset BDD100K show that the VE-Prompt improves the multi-task baseline and further surpasses single-task models.

\let\thefootnote\relax\footnotetext{$^\dagger$Corresponding author.}

\end{abstract}

\section{Introduction}

\begin{figure}[t!]
    \centering
    \includegraphics[width=1.0\linewidth]{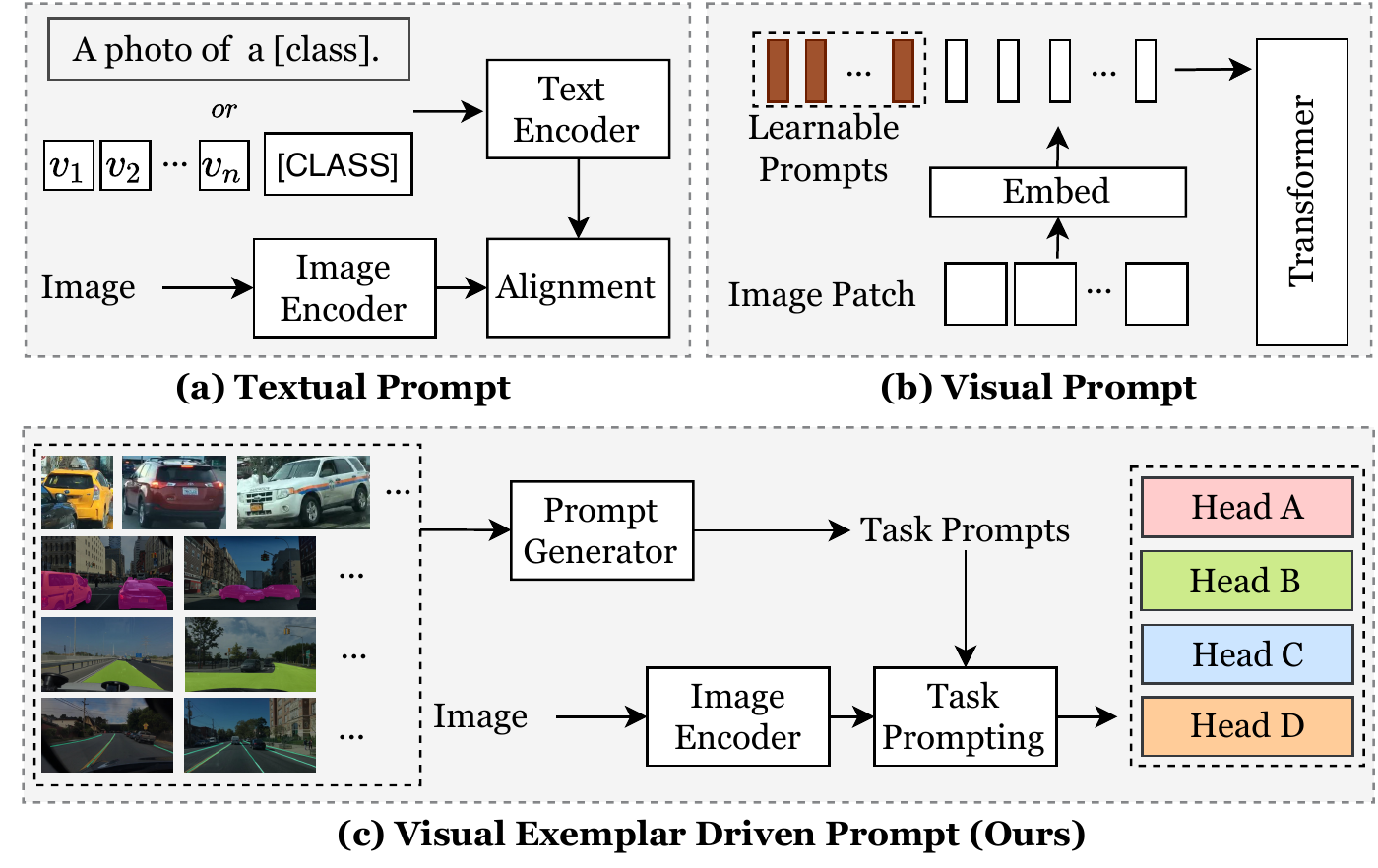}
    \vspace{-4mm}
    \caption{\textbf{Comparison of different prompts in computer vision.} (a) Extracting textual prompts from a text encoder to perform image-text alignment \cite{Zhou2021LearningTP}. (b) Prepend learnable prompts to the embeddings of image patches \cite{jia2022visual}. (c) Visual exemplar driven prompts for multi-task learning (ours). The generated task prompts encode high-quality task-specific knowledge for downstream tasks.}
    \vspace{-4mm}
    \label{fig:prompt}
\end{figure}

Multi-task learning (MTL) has been the source of a number of breakthroughs in autonomous driving over the last few years \cite{kokkinos2017ubernet,Yang2018EndtoendMM,wu2021yolop} and general vision tasks recently \cite{Fifty2021EfficientlyIT,Likhosherstov2021PolyViTCV,marfoq2021federated,bruggemann2021exploring,xu2022multi}.
As the foundation of autonomous driving, a robust vision perception system is required to provide critical information, including the position of traffic participants, traffic signals like lights, signs, lanes, and obstacles that influence the drivable space, to ensure driving safety and comfort.
These tasks gain knowledge from the same data source and present prominent relationships between each other, like traffic participants, are more likely to appear within drivable spaces and traffic signs may appear near traffic lights, etc. Training these tasks independently is time costing and fails to mine the latent relationship among them.
Therefore, it is crucial to solve these multiple tasks simultaneously, which can improve data efficiency and reduce training and inference time.

Some recent works have attempted to apply unified training on multiple tasks in autonomous training. Uncertainty \cite{Kendall2018Multi} trains per-pixel depth prediction, semantic segmentation, and instance segmentation in a single model. CIL \cite{ishihara2021multi} introduces an extra traffic light classifier to learn different traffic patterns following traffic light changes. CP-MTL \cite{chen2018multi} learns object detection and depth prediction together to identify dangerous traffic scenes. However, these works differ in task types, evaluation matrix, and dataset, making it hard to compare their performances. For example, most of them are developed upon dense prediction \cite{bruggemann2021exploring,xu2022multi} and natural language understanding \cite{wang2018glue,clark2019bam}, rather than being tailored for more common perception tasks for autonomous driving, thus these methods may produce poor results when applied to a self-driving system. As a result, there is an emerging demand for a thorough evaluation of existing multi-task learning methods covering common tasks in autonomous driving.

\begin{figure*}[t!]
    \centering
    \includegraphics[width=1.0\linewidth]{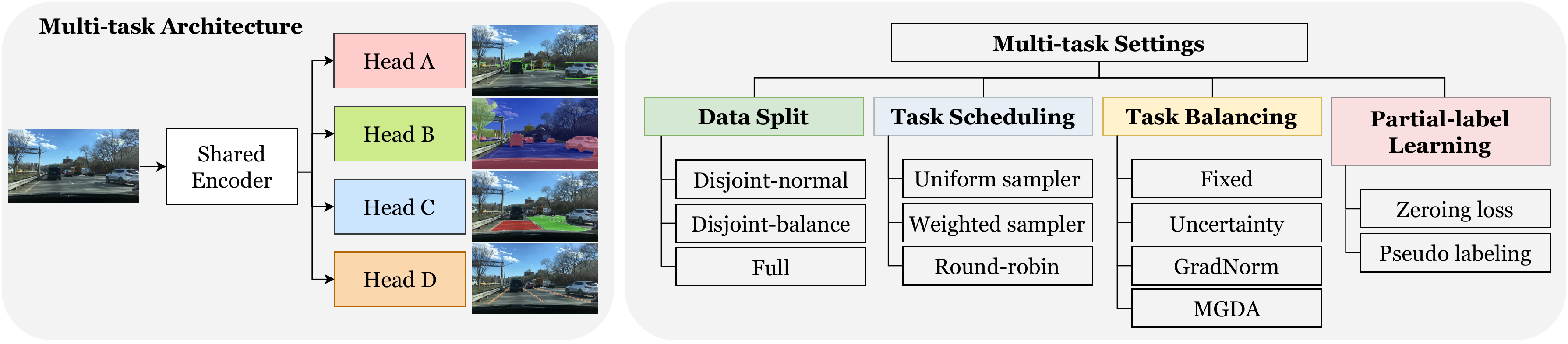}
    \vspace{-4mm}
    \caption{\textbf{The multi-task architecture and settings in our investigation.} We follow the common multi-task architecture where each task shares the same encoder and has its specific head. The multi-task settings focus on three types of task scheduling, four task balancing methods, and two partial-label learning techniques and cover three common data split settings.}
    \vspace{-4mm}
    \label{fig:benchmark}
\end{figure*}

In this paper, we focus on heterogeneous multi-task learning in common scenarios of autonomous driving and cover popular self-driving tasks, i.e., object detection, semantic segmentation, drivable area segmentation, and lane detection.
We provide a systematic study of present MTL methods on large-scale driving dataset BDD100K \cite{yu2020bdd100k}.
Specifically, we find that task scheduling \cite{Likhosherstov2021PolyViTCV} is better than zeroing loss \cite{Xiao2018UnifiedPP}, but worse than pseudo labeling \cite{Ghiasi2021Multi} on most tasks. Interestingly, in task-balancing methods, Uncertainty \cite{Kendall2018Multi} produces satisfactory results on most tasks, while MGDA \cite{sener2018multi} only performs well on lane detection. This indicates that negative transfer \cite{Crawshaw2020MultiTaskLW}, which is a phenomenon that increasing the performance of a model on one task will hurt the performance on another task with different needs, is common among these approaches.

To mitigate the negative transfer problem, we introduce the visual exemplar-driven task-prompting (shorten as \textbf{VE-Prompt}) based on the following motivations:
(1) Given the visual clues of each task, the model can extract task-related information from the pre-trained model.
Different from current prompting methods which introduce textual prompts \cite{Zhou2021LearningTP,zhou2022conditional,rao2022denseclip,chen2022obj2seq} or learnable context \cite{jia2022visual}, we leverage exemplars containing information of target objects to generate  task-specific prompts by considering that the visual clues should represent the specific task to some extent, and give hints for learning task-specific information;
(2) Transformer has achieved competitive performance on many vision tasks but usually requires long training time, thus tackling four tasks simultaneously on a pure transformer is resource-intensive. To overcome this challenge, we efficiently bridge transformer encoders and convolutional layers to build the hybrid multi-task architecture.
Extensive experiments show that VE-Prompt surpasses multi-task baselines by a large margin.

We summarize the main contributions of our work below:
\begin{itemize}[topsep=-2pt,leftmargin=8pt]
    \item We provide an in-depth analysis of current multi-task learning approaches under multiple settings that comply with real-world scenarios, consisting of three common multi-task data split settings, two partial-label learning approaches, three task scheduling techniques, and three task balancing strategies.
    \item We propose an effective framework VE-Prompt, which utilizes visual exemplars to provide task-specific visual clues and guide the model toward learning high-quality task-specific representations.
    \item The VE-Prompt framework is constructed in a computationally efficient way and outperforms competitive multi-task methods on all tasks.
\end{itemize}

\section{Related Work}
\textbf{Multi-task Learning} Multi-task learning jointly trains shared parameters on multiple tasks, mining latent information among them to improve efficiency and accuracy.
Famous multi-task learning models include Mask R-CNN \cite{He2017Mask}, which applies Faster R-CNN \cite{Ren2015Faster} as the backbone and conducts instance segmentation and object detection at the same time.
Other methods like Eigen \emph{et al.}~\cite{Eigen2015Predicting} address depth prediction, surface normal estimation, semantic labeling tasks, and MultiNet~\cite{teichmann2016multinet} provide prediction on classification, detection, semantic segmentation tasks within a single model.
YOLOP~\cite{wu2021yolop} leverages CSPDarknet as the backbone, which branches out three task-specific heads for object detection, drivable area segmentation, and lane detection prediction. 
Standley \emph{et al.} \cite{Standley2020WhichTS} and Christopher \emph{et al.} \cite{Fifty2021EfficientlyIT} improves previous multi-task training schema by grouping proper tasks together rather than naively training all tasks together.
In this paper, we focus on developing general and effective approaches for multi-task learning in autonomous driving scenarios. 

\textbf{Visual Perception for Autonomous Driving} Autonomous driving relies on a perception system to gather information and understand the environment. Visual perception, as the most similar sensing modality to humans, provides high-resolution images that satisfy almost all tasks required for autonomous driving. Some of the tasks have long been studied beyond autonomous driving scenarios. Chen \emph{et al.}~\cite{chen2019learning} predicts 2D object detection from images while Semantic FPN~\cite{kirillov2019panoptic} performs semantic segmentation and Lanenet~\cite{wang2018lanenet} implements lane detection respectively using visual inputs. Though these models are designed for different tasks, they all adopt the backbone-header architecture, some of which even share the same backbone structure like ResNet~\cite{he2016deep} or transformer~\cite{dosovitskiy2020image}. Running independent models for perception tasks separately is a waste of time and computation resources, making an emerging call for the development of a unified perception system.

\textbf{Prompt-based Learning} Prompt-based learning \cite{liu2021pre,he2022hyperprompt,wang2022dualprompt} is put forward to bridge the gap between pre-training and model tuning in the field of natural language processing.
GPT-3 \cite{brown2020language} first designs various text prompts according to the property of tasks and treats the downstream task as a masked language modeling problem. Meanwhile, other approaches like \cite{li2021prefix,liu2021gpt,zhong2021factual} train learnable continuous prompts in the embedding space of the model and achieve competitive performance compared with finetuning.
Recently, CLIP \cite{Radford2021LearningTV}, which is trained on multi-modality vision-language pairs data, achieved impressive performance on zero-shot image classification by injecting visual categories into the text input as prompts. Subsequent works \cite{Zhou2021LearningTP,yao2021cpt,Gao2021CLIPAdapterBV} further tune CLIP with learnable soft prompts by few-shot supervisions in the field of computer vision, or leverage text features from CLIP to enhance visual representations \cite{rao2022denseclip,chen2022obj2seq}. Prompt tuning without textual information is introduced by injecting learnable vectors in the input space \cite{jia2022visual} or inserting lightweight blocks to learn prompts \cite{nie2022pro}.
However, these approaches are tailored for solving downstream tasks independently and are inapplicable to heterogeneous multi-task learning. In this work, we design the visual exemplar-driven task-prompting to inject task-specific knowledge for heterogeneous multi-task learning.

\section{Empirical Study}

\begin{figure*}[h]
    \centering
    \includegraphics[width=1.0\linewidth]{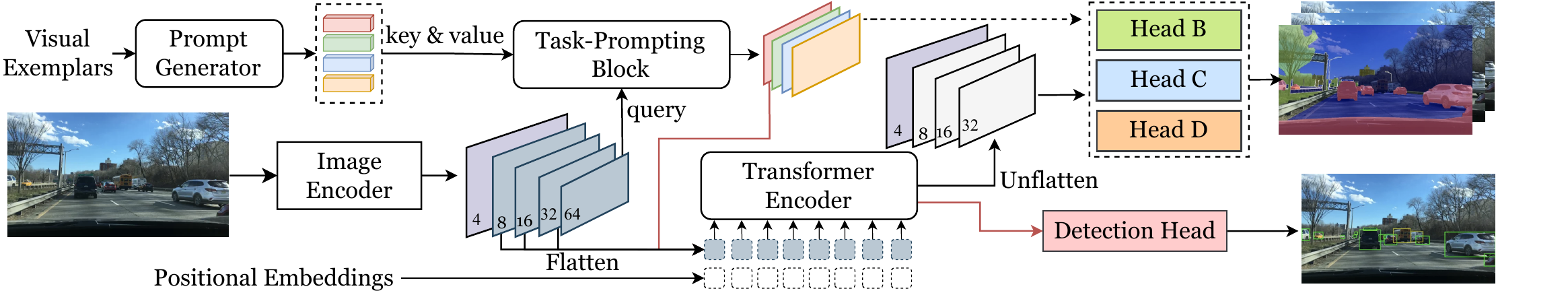}
    \vspace{-4mm}
    \caption{\textbf{The architecture of the proposed VE-Prompt.} VE-Prompt consists of (1) the image encoder to extract image features; (2) a shared transformer encoder for feature enhancement; (3) task-specific prompts generated by the prompt generator with visual exemplars; (4) a task-prompting block to integrate the visual representation with task-specific prompts; and (5) task-specific heads for different tasks.}
    \vspace{-4mm}
    \label{fig:framework}
\end{figure*}

\noindent\textbf{Multi-task Architectures}
Multi-task learning (MTL) architectures apply parameter sharing to learn shared information between different tasks. MTL architectures can be divided into encoder-focused architectures \cite{misra2016cross,ruder2019latent,liu2019end,gao2019nddr} and decoder-focused ones \cite{xu2018pad,zhang2018joint,vandenhende2020mti} according to parameter sharing scope. Encoder-focused architectures can be further categorized into hard and soft parameter sharing. In this paper, we select the hard parameter-sharing structure as our backbone due to its simplicity and stability. Parameters are only shared in the encoder part of the model followed by task-specific heads. As Figure \ref{fig:benchmark} shows, the image inputs first go through the shared encoder, and then the feature map is fed into different heads to produce corresponding predictions.

\noindent\textbf{Task Scheduling} Task scheduling is the process of choosing which task or tasks to train on at each training step. Some scheduling methods arrange the task orders during the training process in a fixed order like Round-Robin~\cite{yu2020bdd100k}, while others may sample tasks following specific distributions~\cite{likhosherstov2021polyvit}, like Uniform sampler and Weighted sampler. 
Specifically, Uniform sampler samples tasks from a uniform distribution and Weighted sampler samples tasks with weight proportional to the number of training epochs of each task.
We test the above three task scheduling methods in our investigation and compare their performances.

\noindent\textbf{Task Balancing} Task balancing is designed to deal with the gradients between tasks for the shared parameters in the network. When dealing with multiple tasks, the shared parameters are likely to be dominated by the one with a large gradient magnitude or confused by conflict gradients. It is intuitive to apply weights over these gradients to balance among tasks, and several methods have been proposed, including 1) Fixed weighting, which fixes all loss weights during training;
2) Uncertainty weighting~\cite{Kendall2018Multi}, introducing the task-dependent Homoscedastic uncertainty as the basis for weighting losses by maximizing the Gaussian likelihood of the uncertainty;
3) GradNorm~\cite{chen2018gradnorm}, calculating the product of $L2$ norm of task gradient and the relative inverse learning rate as the indicator of the task learning pace, and then setting task weights to minimize the learning pace difference among tasks to balance the training process;
4) MGDA \cite{sener2018multi}, treating the Multi-Task Learning problem as a multi-objective optimization problem by using multiple gradient descent algorithm \cite{desideri2012multiple};
5) ParetoMTL \cite{lin2019pareto}, which finds a solution called Pareto optimal solution where all task losses can decrease without increasing the loss on other tasks.

\noindent\textbf{Learning on Partial Labels} Image segmentation task requires annotations of labels to every pixel of the image, which costs a great time, and as a result hard to get enough annotations.
To process the missing annotation problem, two different methods are introduced, including Zeroing loss~\cite{kokkinos2017ubernet,xiao2018unified} and Pseudo labeling~\cite{Ghiasi2021Multi}.
Zeroing loss~\cite{kokkinos2017ubernet,xiao2018unified} simply zero losses for a particular task if the input image does not have the corresponding annotation. 
Pseudo labeling~\cite{Ghiasi2021Multi} first trains a teacher model on fully labeled data. Then the teacher model is used to label the missing annotations to create a multitask pseudo-labeled dataset.

We focus on four major tasks in autonomous driving, i.e., object detection, semantic segmentation, drivable area segmentation, and lane detection. The in-depth analysis of current multi-task methods is shown in Section \ref{sec:multi_task_methods}.

\section{VE-Prompt}
The key to multi-task learning is to learn high-quality task-specific representations among tasks, which can explore relationships between tasks. Therefore, a good multi-task learning framework should take full advantage of task-specific priors, and guide the model to learn better representations.
To this end, we introduce our proposed multi-task framework with VE-Prompt, which consists of five components: (1) an image feature encoder to extract image features; (2) a lightweight shared transformer encoder for feature enhancement; (3) task-specific prompts which encodes task-specific information from visual exemplars; (4) a visual exemplar driven task-prompting block to integrate the visual representation with task-specific prompts; (5) task-specific heads for predicting results simultaneously.

The whole pipeline of the proposed multi-task framework is shown in Figure \ref{fig:framework}. In the following sections, we first delve into the overall multi-task framework in Section \ref{sec:framework} and elaborate on the visual exemplar-driven prompt in Section \ref{sec:prompt_gen} and the task-prompting module in Section \ref{sec:prompting} respectively.

\subsection{Bridging CNN and Transformer}
\label{sec:framework}
The multi-task framework aims to learn more effective representations for all tasks via bridging CNN and Transformer efficiently. The neck of the image encoder and segmentation heads of the framework are CNN-based, reducing the overall training time. The shared transformer encoder is built upon the transformer architecture to capture the long-range dependency \cite{xie2021cotr}.

\noindent\textbf{Image Encoder}
The image encoder consists of a backbone network and a neck network. We choose the Swin transformer \cite{liu2021swin} as the backbone to extract features of the input image. The output of the backbone is denoted as $\{ C_2, C_3, C_4, C_5 \}$. Then we adopt Feature Pyramid Network (FPN) \cite{lin2017feature} module for the neck network to fuse features generated by the backbone. The pyramidal features are of 5 scales, and the detection head only processes the last four-scale features for reducing the computation cost. Here we denote the output of the neck as $\{ P_2, P_3, P_4, P_5, P_6 \}$, which have strides of \{4, 8, 16, 32, 64\} pixels.

\noindent\textbf{Shared Transformer Encoder}
The shared transformer encoder TransEncoder receives multi-scale outputs from the neck and enhances features for following task-specific heads. We first flatten the feature maps from $\{ P_3, P_4, P_5, P_6 \}$ and concatenate them to obtain a 1D sequence $P$. Since flattening the features leads to losing the spatial information critical for segmentation, we supplement positional embeddings $p_l$ to the flattened features. For the model not considering prompts, we obtain the enhanced feature as follows:
\begin{equation}
    O = \text{TransEncoder}(P + p_l).
\label{equ:trans_encoder}
\end{equation}
After feature enhancement, $O$ is passed to the detection head directly, while unflattened to multi-scale features $\{ z_3, z_4, z_5, z_6 \}$ for segmentation heads.

\noindent\textbf{Detection Head}
The detection head consists of 4 multi-scale deformable decoder layers which are elaborated in DINO \cite{zhang2022dino}.
Following DINO, we adopt the mixed query selection strategy to initialize anchors as positional queries for the decoder and use the contrastive denoising training approach by taking into account hard negative samples.

\noindent\textbf{Segmentation Head}
For segmentation-based tasks, we choose Semantic FPN \cite{kirillov2019panoptic} as the segmentation head. In the model without considering prompts, segmentation heads take in multi-layer features from both the neck and shared transformer encoder $\{ P_2, z_3, z_4, z_5 \}$. The resolution of $P_2$ is larger and thus provides more image information for the following heads. Then the multi-layer features are up-sampled and summed element-wisely. This merged feature map is again upsampled $4\times$ followed by softmax to produce the classification score for every pixel at the original resolution.

\subsection{Prompt Generation with Visual Exemplar}
\label{sec:prompt_gen}

\begin{figure}[t!]
    \centering
    \includegraphics[width=1.0\linewidth]{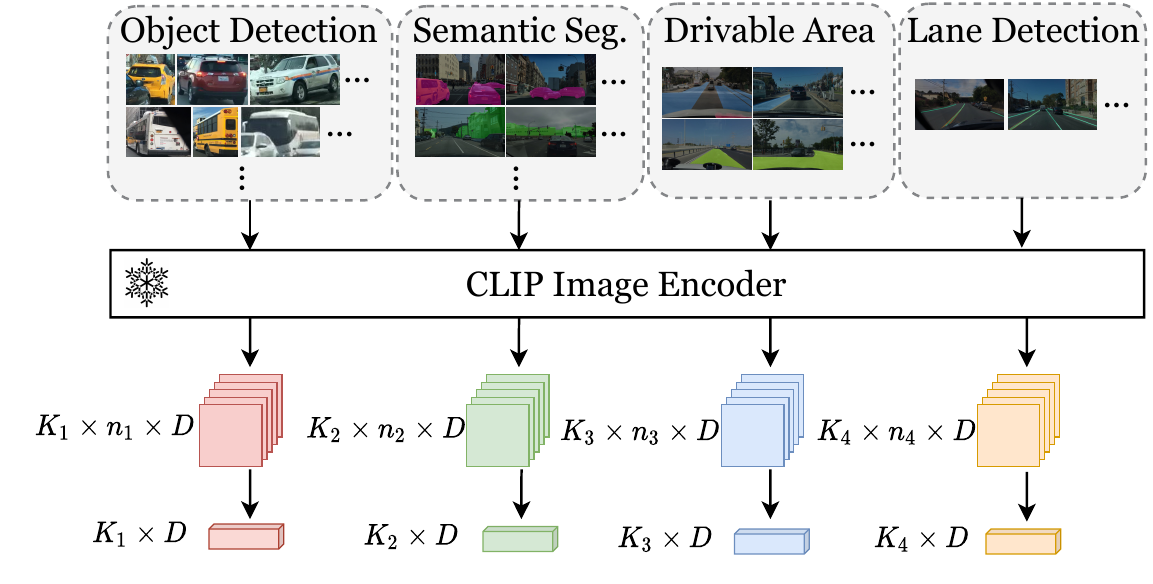}
    \vspace{-6mm}
    \caption{\textbf{Process of generating visual exemplar-driven prompts.} For the box-wise task, we crop class-related image regions to generate visual exemplars. For pixel-wise tasks, we mask class-related image regions with colored segmentation masks to produce exemplars. Then the fixed CLIP image encoder is adopted to extract task-specific prompts.}
    \vspace{-6mm}
    \label{fig:prompt}
\end{figure}

In order to motivate the model to learn more high-quality task-specific knowledge and handle all tasks better, VE-Prompt is introduced to provide more task-specific information with visual clues.
The process of generating visual exemplar-driven prompts is shown in Figure \ref{fig:prompt}.
The key idea of task-specific prompts is to let the model know how to solve different tasks and what categories to focus on in advance of each task. Therefore, task-specific prompts should contain object-level information which helps the model understand tasks better, and we leverage visual exemplars to generate prompts.

We first sample a few examples from the training set to generate object-level image regions and segmentation masks as in Figure \ref{fig:prompt}. There are a few generated exemplars for each category in each task, thus the generated prompts cover all classes.
Then we adopt CLIP \cite{Radford2021LearningTV} to generate task-specific prompts since it is a robust feature extractor pre-trained with a huge amount of image-text data pairs. For visual perception, the ground-truth annotations can provide hints of the shapes and sizes of different objects, motivating the model to learn high-quality task-specific representations.
The task-specific prompt is object-level for object detection and aims to represent relevant image regions.
Only the generated prompts are used during training and inference, and no new exemplars will be included, thus it will not lead to training data leakage.

For the box-wise task (object detection), we use the annotated bounding boxes to crop sampled images and obtain raw object-level image regions for generating prompts. We choose the image encoder with ViT \cite{dosovitskiy2020image} backbone and pass $n$ image regions $\{r_i^k\}$ of $K$ classes to get the initial prompt as follows:
\begin{equation}
\begin{aligned}
    \hat{\{p_i^k\}} = \text{L2\_NORM}(\text{IE}(\{r_i^k\})) \in \mathbb{R}^{K \times n \times D}, \\
    p = \frac{1}{n}\sum\limits_{i}^{n} \hat{\{p_i^k\}} \in \mathbb{R}^{K \times D}, i = 1, 2, ..., n,
\end{aligned}
\label{equ:prompt}
\end{equation}
where IE and $D$ represent the image encoder of CLIP and the feature dimension respectively. $n$ stands for the number of visual exemplars for each category. $\hat{\{p_i^k\}}$ and $p$ indicate all prompts from image regions and the averaged version respectively. Specifically, class numbers for detection, semantic segmentation, drivable area segmentation, and lane detection are denoted as $K_1$, $K_2$, $K_3$, and $K_4$. The number of visual exemplars for different tasks is further denoted as $n_1$, $n_2$, $n_3$, and $n_4$.

For pixel-wise tasks (i.e., semantic segmentation, drivable area segmentation, and lane detection), image regions of specific classes are marked with colored segmentation masks, and different colors indicate different object categories of different tasks. Similar to object detection, after obtaining $n$ images with colored segmentation masks of $K$ classes, we adopt CLIP to extract features by Equation \ref{equ:prompt}.
In this way, we get task-specific prompts $p_{\text{det}}$, $p_{\text{sem}}$, $p_{\text{driv}}$, $p_{\text{lane}}$ for object detection, semantic segmentation, drivable area segmentation, and lane detection, respectively.

\subsection{Visual Exemplar Driven Task Prompting}
\label{sec:prompting}
Task prompting aims to integrate the image features with task-specific prompts to obtain high-quality task-specific representations.
It receives additional task-specific prompts as inputs and generates task-specific features for following task-specific heads.

Here we design two prompting methods to improve task-specific representations.
The first strategy is pre-head prompting. The last feature map $P_6$ from the neck and task-specific prompt $p$ are fused via a transformer decoder:
\begin{equation}
    f_{pre} = \text{TransDecoder}(q=P_6, k=p, v=p),
\end{equation}
where $q$, $k$ and $v$ stand for query, key and value. In this way, we get task-specific features $f_{pre}^{det}$, $f_{pre}^{sem}$, $f_{pre}^{driv}$, and $f_{pre}^{lane}$ for object detection, semantic segmentation, drivable area segmentation, and lane detection, respectively. For object detection, we flatten $\{  P_3, P_4, P_5, f_{pre}^{det} \}$ to a 1D sequence and combine it with positional embeddings. Following Equation \ref{equ:trans_encoder}, we obtain features for the detection head $O_{det}$.
For segmentation-based tasks, we first flatten $\{  P_3, P_4, P_5 \}$ to a 1D sequence and get the enhanced features through Equation \ref{equ:trans_encoder}.
Then unflatten the output features as $\{ z_3', z_4', z_5' \}$, and pass $\{ P_2, z_3', z_4', z_5', f_{pre}]$ to specific segmentation-based heads. Note that $f_{pre}$ is marked as $f_{pre}^{sem}$, $f_{pre}^{driv}$, or $f_{pre}^{lane}$ according to the task type.

Another choice is to refine predicted results with task-specific prompts, namely post-head prompting. In this variant, we obtain class-related features ($\mathbb{R}^{K \times D}$) similar as follows:
\begin{equation}
    f_{post} = \text{TransDecoder}(q=p, k=P_6, v=P_6).
\end{equation}
Here class-related outputs from task-specific heads are denoted as $v$. Then the final output is calculated as:
\begin{equation}
    v' = \text{MLP}(v \cdot f_{post}).
\end{equation}

Empirical results of these two strategies are presented in Section \ref{sec:ablation} and show that pre-head prompting performs better than post-head prompting.

\begin{table*}[t]
    \centering
    \caption{Comparisons of popular task scheduling strategies and partial-label learning methods.}
    \vspace{-2mm}
    \resizebox{1.0\linewidth}{!}{
    \begin{tabular}{c|c|ccc|c|c|c|c|c}
        \toprule
        Setting & Methods & mAP & AP50 & AP75 & mIoU (SS) & mIoU (DA) & IoU (LD) & Avg. & $\Delta_{MTL}(\%)$ \\
        \midrule
        \multirow{3}{*}{Full} & Zeroing loss \cite{Xiao2018UnifiedPP} & 36.2 & 61.6 & 35.9 & 58.6 & 89.3 & 23.8 & 52.0 & -2.68 \\
         & Pseudo labeling \cite{Ghiasi2021Multi} & 36.3 & 61.6 & 36.1 & 60.9 & 89.3 & 23.8 & 52.6 & -1.65 \\
        \cmidrule{2-10}
         & VE-Prompt (Ours) & \tfbestfirst{39.2} & \tfbestfirst{64.9} & \tfbestfirst{39.0} & \tfbestfirst{63.2} & \tfbestfirst{89.4} & \tfbestfirst{24.0} & \tfbestfirst{54.0} & \tfbestfirst{+1.52} \\
        \midrule
        \multirow{6}{*}{Disjoint-normal} & Zeroing loss \cite{Xiao2018UnifiedPP} & 31.1 & 54.3 & 30.2 & 55.7 & 88.0 & 22.2 & 49.3 & -2.64 \\
         & Uniform sampler \cite{Likhosherstov2021PolyViTCV} & 30.1 & 52.8 & 29.0 & 60.6 & 88.6 & 23.4 & 50.7 & -0.10 \\
         & Weighted sampler \cite{Likhosherstov2021PolyViTCV} & 29.3 & 51.9 & 28.7 & 58.5 & \tfbestfirst{88.9} & \tfbestfirst{23.8} & 50.1 & -1.19 \\
         & Round-robin \cite{Likhosherstov2021PolyViTCV} & 30.2 & 53.1 & 29.7 & 61.0 & 88.7 & 23.5 & 50.9 & +2.87 \\
         & Pseudo labeling \cite{Ghiasi2021Multi} & 32.6 & 54.6 & 32.3 & 59.7 & 88.2 & 23.0 & 50.9 & +1.19 \\
        \cmidrule{2-10}
         & VE-Prompt (Ours) & \tfbestfirst{34.2} & \tfbestfirst{56.9} & \tfbestfirst{33.9} & \tfbestfirst{62.2} & 88.3 & 23.3 & \tfbestfirst{52.0} & \tfbestfirst{+3.95} \\
        \midrule
        \multirow{5}{*}{Disjoint-balance} & Zeroing loss \cite{Xiao2018UnifiedPP} & 29.7 & 52.3 & 29.2 & 57.5 & 86.7 & 21.4 & 48.8 & -1.61 \\
         & Uniform sampler \cite{Likhosherstov2021PolyViTCV} & 28.1 & 50.2 & 27.5 & 60.4 & 87.1 & \tfbestfirst{22.6} & 50.0 & -0.44 \\
         & Round-robin \cite{Likhosherstov2021PolyViTCV} & 28.4 & 50.8 & 27.8 & 60.0 & 87.1 & \tfbestfirst{22.6} & 49.5 & -0.34 \\
         & Pseudo labeling \cite{Ghiasi2021Multi} & 31.3 & 52.8 & 30.8 & 60.2 & 87.0 & 22.2 & 50.2 & +1.87 \\
        \cmidrule{2-10}
         & VE-Prompt (Ours) & \tfbestfirst{33.9} & \tfbestfirst{56.6} & \tfbestfirst{33.7} & \tfbestfirst{61.2} & \tfbestfirst{87.4} & 22.2 & \tfbestfirst{51.2} & \tfbestfirst{+4.72} \\
        \bottomrule
    \end{tabular}}
    \vspace{-2mm}
    \label{tab:multi_task_baselines}
\end{table*}

\begin{table*}[t]
    \centering
    \caption{Comparisons of task balancing strategies with pseudo labels. $^*$ means using the full image encoder to compute the gradient norm.}
    \vspace{-2mm}
    \resizebox{1.0\linewidth}{!}{
    \begin{tabular}{c|c|ccc|c|c|c|c|c}
        \toprule
        Setting & Method & mAP & AP50 & AP75 & mIoU (SS) & mIoU (DA) & IoU (LD) & Avg. & $\Delta_{MTL}(\%)$ \\
        \midrule
        \multirow{4}{*}{Full} & Fixed \cite{Ghiasi2021Multi} & 36.3 & 61.6 & 36.1 & 60.9 & 89.3 & 23.8 & 52.6 & -1.65 \\
         & Uncertainty \cite{Kendall2018Multi} & 36.2 & 61.6 & 35.5 & 61.2 & \tfbestfirst{89.5} & \tfbestfirst{24.6} & 52.9 & -0.76 \\
         & GradNorm \cite{chen2018gradnorm} & 23.4 & 40.9 & 22.8 & 25.8 & 51.3 & 13.0 & 28.4 & -46.24  \\
        \cmidrule{2-10}
         & VE-Prompt (Ours) & \tfbestfirst{39.2} & \tfbestfirst{64.9} & \tfbestfirst{39.0} & \tfbestfirst{63.2} & 89.4 & 24.0 & \tfbestfirst{54.0} & \tfbestfirst{+1.52} \\
        \midrule
        \multirow{5}{*}{Disjoint-normal} & Fixed \cite{Ghiasi2021Multi} & 32.6 & 54.6 & 32.3 & 59.7 & 88.2 & 23.0 & 50.9 & +1.19 \\
         & Uncertainty \cite{Kendall2018Multi} & 32.2 & 54.1 & 31.5 & 59.8 & \tfbestfirst{88.6} & 23.8 & 51.1 & +1.79 \\
         & GradNorm \cite{chen2018gradnorm} & 25.9 & 43.2 & 26.1 & 39.2 & 39.6 & 3.7 & 27.1 & -46.18 \\
         & MGDA \cite{sener2018multi} & 25.9 & 44.6 & 26.0 & 50.1 & 85.4 & \tfbestfirst{25.2} & 46.7 & -7.26 \\
        \cmidrule{2-10}
         & VE-Prompt (Ours) & \tfbestfirst{34.2} & \tfbestfirst{56.9} & \tfbestfirst{33.9} & \tfbestfirst{62.2} & 88.3 & 23.3 & \tfbestfirst{52.0} & \tfbestfirst{+3.95} \\
        \midrule
        \multirow{5}{*}{Disjoint-balance} & Fixed \cite{Ghiasi2021Multi} & 31.3 & 52.8 & 30.8 & 60.2 & 87.0 & 22.2 & 50.2 & +1.87 \\
         & Uncertainty \cite{Kendall2018Multi} & 31.2 & 53.1 & 30.9 & 59.9 & 87.0 & 22.2 & 50.1 & +1.66 \\
         & GradNorm \cite{chen2018gradnorm} & 28.9 & 49.0 & 28.7 & 46.8 & 57.4 & 19.6& 38.2 & -17.26 \\
         & GradNorm$^*$ \cite{chen2018gradnorm} & 30.7 & 51.8 & 30.4 & 56.6 & 86.9 & 21.7 & 49.0 & -0.73 \\
         & MGDA \cite{sener2018multi} & 21.0 & 38.0 & 20.3 & 45.5 & 82.7 & \tfbestfirst{24.3} & 43.4 & -12.48 \\
        \cmidrule{2-10}
         & VE-Prompt (Ours) & \tfbestfirst{33.9} & \tfbestfirst{56.6} & \tfbestfirst{33.7} & \tfbestfirst{61.2} & \tfbestfirst{87.4} & 22.2 & \tfbestfirst{51.2} & \tfbestfirst{+4.72} \\
        \bottomrule
    \end{tabular}}
    \vspace{-3mm}
    \label{tab:multi_task_optimize}
\end{table*}

\subsection{Optimization}
Since there are four different perception tasks in our network, our multi-task loss contains four parts. For object detection, we adopt the same objective as in DINO \cite{zhang2022dino}. For segmentation-based tasks, i.e., semantic segmentation, drivable area segmentation, and lane detection, we employ the same loss as in Semantic FPN \cite{kirillov2019panoptic}. Therefore, the total loss for the multi-task model is formulated as follows:
\begin{equation}
    \mathcal{L}_{\text{total}} = \lambda_{det}\mathcal{L}_{\text{det}} + \lambda_{sem}\mathcal{L}_{\text{sem}} + \lambda_{driv}\mathcal{L}_{\text{driv}} + \lambda_{lane}\mathcal{L}_{\text{lane}},
\end{equation}
where $\mathcal{L}_{\text{det}}$, $\mathcal{L}_{\text{sem}}$, $\mathcal{L}_{\text{driv}}$, $\mathcal{L}_{\text{lane}}$ represent objectives for object detection, semantic segmentation, drivable area segmentation, and lane detection, respectively. $\lambda_{det}$, $\lambda_{sem}$, $\lambda_{driv}$ and $\lambda_{lane}$ stand for different loss weights.

\section{Experiments}
\subsection{Dataset Settings}
\label{sec:data_split}
Our experiments are tested on the BDD100K dataset. BDD100K dataset has $\sim$74k training images and covers both object detection (OD), semantic segmentation (SS), drivable area segmentation (DA), and lane detection (LD).
We follow \cite{liang2022effective} to consider three dataset split settings complying with real-world scenarios, i.e., Disjoint-normal setting, Disjoint-balance setting, and Full setting:

\noindent\textbf{Disjoint-normal Setting} 
The number of labeled images for each task is as follows: object detection (10k), semantic segmentation (7k), drivable area segmentation (20k), and lane detection (20k).

\noindent\textbf{Disjoint-balance Setting} 
There are 21k images in this set and each task has 7k labeled images that are not overlapped with other tasks.

\noindent\textbf{Full Setting} 
Full-setting refers to experimenting on all available annotations on $\sim$74k images in BDD100K and can be used to analyze the upper bound of different methods.

\begin{table*}[t]
    \centering
    \caption{Comparison between single-task and multi-task learning baselines under different settings.}
    \vspace{-2mm}
    \resizebox{1.0\linewidth}{!}{
    \begin{tabular}{c|c|ccc|c|c|c|c|c}
        \toprule
        Setting & Methods & mAP & AP50 & AP75 & mIoU (SS) & mIoU (DA) & IoU (LD) & Avg. & $\Delta_{MTL} (\%)$ \\
        \midrule
        \multirow{8}{*}{Full} & Sparse R-CNN \cite{Sun2021Sparse} & 36.5 & 61.5 & 36.1 & - & - & - & - & - \\
         & DINO \cite{zhang2022dino} & 38.6 & 64.2 & 38.2 & - & - & - & - & - \\
         & Semantic FPN \cite{kirillov2019panoptic} & - & - & - & 59.8 & - & - & - & - \\
         & Semantic FPN \cite{kirillov2019panoptic} & - & - & - & - & 89.1 & - & - & - \\
         & Semantic FPN \cite{kirillov2019panoptic} & - & - & - & - & - & 25.9 & - & - \\
        \cmidrule{2-10}
         & Sparse R-CNN based & 36.3 & 61.6 & 36.1 & 60.9 & 89.3 & 23.8 & 52.6 & -1.65 \\
         & DINO based & \tfbestfirst{39.4} & 64.5 & \tfbestfirst{39.8} & 61.5 & 84.9 & 22.0 & 52.0 & -2.25 \\
         & VE-Prompt (Ours) & 39.2 & \tfbestfirst{64.9} & 39.0 & \tfbestfirst{63.2} & \tfbestfirst{89.4} & \tfbestfirst{24.0} & \tfbestfirst{54.0} & \tfbestfirst{+1.52} \\
        \midrule
        \multirow{8}{*}{Disjoint-normal} & Sparse R-CNN \cite{Sun2021Sparse} & 28.8 & 50.4 & 28.0 & - & - & - & - & - \\
         & DINO \cite{zhang2022dino} & 31.2 & 53.0 & 30.5 & - & - & - & - & - \\
         & Semantic FPN \cite{kirillov2019panoptic} & - & - & - & 59.8 & - & - & - & - \\
         & Semantic FPN \cite{kirillov2019panoptic} & - & - & - & - & 87.8 & - & - & - \\
         & Semantic FPN \cite{kirillov2019panoptic} & - & - & - & - & - & 25.2 & - & - \\
        \cmidrule{2-10}
         & Sparse R-CNN based & 32.6 & 54.6 & 32.3 & 59.7 & 88.2 & 23.0 & 50.9 & +1.19 \\
         & DINO based & 33.1 & 55.9 & 32.2 & 59.2 & 87.2 & 22.7 & 50.6 & +0.83 \\
         & VE-Prompt (Ours) & \tfbestfirst{34.2} & \tfbestfirst{56.9} & \tfbestfirst{33.9} & \tfbestfirst{62.2} & \tfbestfirst{88.3} & \tfbestfirst{23.3} & \tfbestfirst{52.0} & \tfbestfirst{+3.95} \\
        \midrule
        \multirow{8}{*}{Disjoint-balance} & Sparse R-CNN \cite{Sun2021Sparse} & 28.1 & 49.2 & 26.7 & - & - & - & - & - \\
         & DINO \cite{zhang2022dino} & 29.4 & 50.8 & 28.1 & - & - & - & - & - \\
         & Semantic FPN \cite{kirillov2019panoptic} & - & - & - & 59.8 & - & - & - & - \\
         & Semantic FPN \cite{kirillov2019panoptic} & - & - & - & - & 85.5 & - & - & - \\
         & Semantic FPN \cite{kirillov2019panoptic} & - & - & - & - & - & 23.7 & - & - \\
        \cmidrule{2-10}
         & Sparse R-CNN based & 31.3 & 52.8 & 30.8 & 60.2 & 87.0 & \tfbestfirst{22.2} & 50.2 & +1.87 \\
         & DINO based & 33.5 & 55.6 & 33.1 & 58.1 & 85.2 & 21.4 & 50.0 & +1.58 \\
         & VE-Prompt (Ours) & \tfbestfirst{33.9} & \tfbestfirst{56.6} & \tfbestfirst{33.7} & \tfbestfirst{61.2} & \tfbestfirst{87.4} & \tfbestfirst{22.2} & \tfbestfirst{51.2} & \tfbestfirst{+4.72} \\
        \bottomrule
    \end{tabular}}
    \vspace{-4mm}
    \label{tab:compare}
\end{table*}

\subsection{Evaluation and Implementation Details}
\noindent\textbf{Evaluation Metric}
In addition to reporting performance on every individual task, we follow \cite{vandenhende2021multi} to evaluate the whole multi-task performance:
\begin{equation}
    \Delta_{MTL} = \frac{1}{T}\sum\limits_{i}^{T}(M_{m,i} - M_{b,i})/M_{b,i},
\end{equation}
where $M_{m,i}$ is the performance of multi-task model on task $i$, and $M_{b,i}$ indicates the result of single-task baseline. Since we choose Sparse R-CNN as the detection head for re-implementing current multi-task methods, we regard it as the baseline for object detection. For object detection, we adopt mAP as the main metric. While for segmentation tasks, we use mIoU to evaluate the model.
We also compute the average performance (Avg.) of all tasks to compare experimental results more intuitively.

\noindent\textbf{Implementation Details}
The default training setting is that epoch and batch size are fixed as 36 and 16, the learning rate is set to $1\times10^{-5}$, and weight decay is $1\times10^{-4}$. We adopt the AdamW optimizer, for which the warmup length is 1 epoch and the warmup factor is 0.001. We choose Swin-Tiny \cite{liu2021swin} as the backbone by default. More details are provided in Appendix.

\subsection{Comparison of Multi-task Methods}
\label{sec:multi_task_methods}
We study the performances of popular existing multi-task methods under three settings on BDD100K. We adopt Sparse R-CNN to construct the detection head for efficiency.

\noindent\textbf{Partial-label Learning}
As shown in Table \ref{tab:multi_task_baselines}, pseudo labeling \cite{Ghiasi2021Multi} can improve performances, especially in object detection and semantic segmentation compared with zeroing loss \cite{Xiao2018UnifiedPP}. Pseudo-labeling achieves satisfactory performance in all settings.

\noindent\textbf{Task Scheduling}
As shown in Table \ref{tab:multi_task_baselines}, three task sampling methods (i.e., Uniform sampler \cite{Likhosherstov2021PolyViTCV}, Weighted sampler \cite{Likhosherstov2021PolyViTCV} and Round-robin \cite{Likhosherstov2021PolyViTCV}) perform better than Zeroing loss \cite{Xiao2018UnifiedPP} by a large margin on segmentation-based tasks, but get worse in object detection. We hypothesize that training one task per step may lead to forgetting to some extent.

\noindent\textbf{Task Balancing}
We choose pseudo labeling as the baseline since task-balancing methods are more suitable in settings with complete labels. Fixed denotes fixed loss weights for all tasks during training. As shown in Table \ref{tab:multi_task_optimize}, Uncertainty performs better than Fixed on semantic segmentation and drivable area segmentation under the full and disjoint-normal settings, while performances of other approaches (i.e., GradNorm and MGDA) degrade significantly. Especially, GradNorm uses the last shared layer of weights to compute gradient norm in its paper, thus we adopt the last layer of $P_5$ in the neck. When we use the full image encoder to compute the gradient norm, GradNorm achieves much better results but still lags behind the baseline. Interestingly, MGDA achieves the best result on lane detection, indicating that it suffers from heavy negative transfer.

For efficiency and effectiveness, we choose pseudo labeling with fixed loss weights as our baseline, which achieves competitive performance compared with other complicated multi-task methods, to verify the effectiveness of VE-Prompt.

\subsection{Compare VE-Prompt with Previous Methods}
As shown in Table \ref{tab:compare}, our VE-Prompt surpasses the baseline consistently on almost all metrics in all three settings and achieves significant overall multi-task performance.
We conclude that VE-Prompt can learn high-quality task-specific knowledge during training, and further improve performance.
VE-Prompt also achieves the best results on three tasks compared with single-task models.
We also compare VE-Prompt with LV-Adapter \cite{liang2022effective} in Appendix.

\begin{table}[t]
    \centering
    \caption{Ablation study of modules in our proposed VE-Prompt. TE means transformer encoder.}
    \vspace{-2mm}
    \resizebox{1.0\linewidth}{!}{
    \begin{tabular}{c|c|c|c|c}
        \toprule
         & mAP & mIoU (SS) & mIoU (DA) & IoU (LD) \\
        \midrule
        DINO based & 33.5 & 58.1 & 85.2 & 21.4 \\
        w/ shared TE & 32.2 & 60.5 & 86.5 & 21.4 \\
        $+$ Prompt & \textbf{33.9} & \textbf{61.2} & \textbf{87.4} & \textbf{22.2} \\
        \bottomrule
    \end{tabular}}
    \vspace{-4mm}
    \label{tab:module}
\end{table}

\subsection{Ablation Study}
\label{sec:ablation}
We conduct all ablation studies under the disjoint-balance setting for efficiency.

\begin{table}[t]
    \centering
    \caption{Ablation study of task-specific prompts. Post and Pre indicate post-head prompting and pre-head prompting respectively.}
    \vspace{-2mm}
    \resizebox{1.0\linewidth}{!}{
    \begin{tabular}{c|ccc|c|c|c}
        \toprule
        \# & Prompt & Post & Pre & mAP & mIoU (SS) & mIoU (DA) \\
        \midrule
        1 & \xmark & \xmark & \xmark & 32.2 & 60.5 & 86.5 \\
        \midrule
        2 & \colorbox{cyan!10}\cmark & \colorbox{cyan!10}\cmark & \xmark & 33.2 & 58.9 & 86.4 \\
        3 & \colorbox{cyan!10}\cmark & \xmark & \colorbox{cyan!10}\cmark & \textbf{33.9} & \textbf{61.2} & \textbf{87.4} \\
        \bottomrule
    \end{tabular}}
    \vspace{-4mm}
    \label{tab:fusion_design}
\end{table}

\noindent\textbf{Module Components}
We present detailed comparisons on each module to validate our VE-Prompt as in Table \ref{tab:module}.
The introduced shared transformer encoder alleviates this imbalance to some extent (row 1 vs. row 2).
Equipped with task-specific prompts through task prompting, the model gets better results on all tasks (row 2 vs. row 3), confirming that task-specific prompts can motivate the model to learn useful task-specific knowledge for specific tasks.

\begin{table}[t]
    \centering
    \caption{Ablation study of initialization for prompt vectors.}
    \vspace{-2mm}
    \resizebox{1.0\linewidth}{!}{
    \begin{tabular}{c|c|c|c|c}
        \toprule
         CLIP Initialization & mAP & mIoU (SS) & mIoU (DA) & IoU (LD) \\
        \midrule
         \xmark & 33.5 & 61.0 & 87.2 & 21.9 \\
         \colorbox{cyan!10}\cmark & \textbf{33.9} & \textbf{61.2} & \textbf{87.4} & \textbf{22.2} \\
        \bottomrule
    \end{tabular}}
     \vspace{-4mm}
    \label{tab:prompt}
\end{table}

\noindent\textbf{Task Prompting}
We conduct an ablation study to compare the proposed two prompting strategies as in Table \ref{tab:fusion_design}. We can see that both post-head and pre-head prompting improve the performance of object detection. However, post-head prompting gets inferior results on segmentation-based tasks (\#1 vs. \#2), indicating that the post-head process is not suitable for dense prediction tasks. On the contrary, pre-head prompting helps the model make full use of task-specific knowledge and improves all tasks consistently (\#3).

\noindent\textbf{Prompt Initialization}
The task-specific prompts are initialized with the pre-trained image encoder of CLIP. We compare it with random initialization as in Table \ref{tab:prompt}. Results show that prompts with CLIP initialization improve the multi-task model on all metrics. More comparisons and analyses on prompting are presented in Appendix.

\section{Conclusion and Discussion}
In this paper, we first provide an in-depth analysis of popular multi-task learning methods under the realistic scenarios of self-driving, which covers four common perception tasks, i.e., object detection, semantic segmentation, drivable area segmentation, and lane detection. We find that existing methods cannot solve all tasks satisfactorily due to the negative transfer. To mitigate the negative transfer, we propose visual exemplar driven task-prompting (VE-Prompt), which incorporates visual exemplars of different tasks to provide high-quality task-specific knowledge. Besides, the proposed framework bridges transformer and convolutional layers for efficient and accurate unified perception in autonomous driving. Experimental results show that VE-Prompt can achieve superior performance on large-scale driving dataset BDD100K.

\noindent\textbf{Limitations}
Although our method has achieved substantial improvement in the overall multi-task metric, we find the model gets worse results on lane detection compared with the single-task model. We conjecture that it is because the lane detection task is quite different from other tasks, thus it is difficult for the multi-task model to solve all four tasks satisfactorily. Our VE-Prompt gets better results on lane detection compared with multi-task baselines, but there is still room for improvement. We believe VE-Prompt can be further improved by introducing more robust pseudo-labeling methods or designing specific heads for lane detection. The proposed prompt generation with visual exemplar is general and can be applied to other applications in computer vision. Our method does not directly involve societal issues.

\section*{Acknowledgements}
We gratefully acknowledge the support of MindSpore\footnote{\url{https://www.mindspore.cn/}}, CANN~(Computer Architecture for Neural Networks) and Ascend AI Processor used for this research.

{\small
\bibliographystyle{ieee_fullname}
\bibliography{main}

\begin{thebibliography}{10}\itemsep=-1pt

\bibitem{brown2020language}
Tom~B. {Brown}, Benjamin {Mann}, Nick {Ryder}, Melanie {Subbiah}, Jared
  {Kaplan}, Prafulla {Dhariwal}, Arvind {Neelakantan}, Pranav {Shyam}, Girish
  {Sastry}, Amanda {Askell}, Sandhini {Agarwal}, Ariel {Herbert-Voss}, Gretchen
  {Krueger}, Tom {Henighan}, Rewon {Child}, Aditya {Ramesh}, Daniel~M.
  {Ziegler}, Jeffrey {Wu}, Clemens {Winter}, Christopher {Hesse}, Mark {Chen},
  Eric {Sigler}, Mateusz {Litwin}, Scott {Gray}, Benjamin {Chess}, Jack
  {Clark}, Christopher {Berner}, Sam {McCandlish}, Alec {Radford}, Ilya
  {Sutskever}, and Dario {Amodei}.
\newblock Language models are few-shot learners.
\newblock In {\em Advances in Neural Information Processing Systems (NeurIPS)},
  volume~33, pages 1877--1901, 2020.

\bibitem{bruggemann2021exploring}
David Br{\"u}ggemann, Menelaos Kanakis, Anton Obukhov, Stamatios Georgoulis,
  and Luc Van~Gool.
\newblock Exploring relational context for multi-task dense prediction.
\newblock In {\em Proceedings of the IEEE/CVF International Conference on
  Computer Vision}, pages 15869--15878, 2021.

\bibitem{chen2019learning}
Rui Chen, Haizhou Ai, Chong Shang, Long Chen, and Zijie Zhuang.
\newblock Learning lightweight pedestrian detector with hierarchical knowledge
  distillation.
\newblock In {\em 2019 IEEE International Conference on Image Processing
  (ICIP)}, pages 1645--1649. IEEE, 2019.

\bibitem{chen2018multi}
Yaran Chen, Dongbin Zhao, Le Lv, and Qichao Zhang.
\newblock Multi-task learning for dangerous object detection in autonomous
  driving.
\newblock {\em Information Sciences}, 432:559--571, 2018.

\bibitem{chen2018gradnorm}
Zhao Chen, Vijay Badrinarayanan, Chen-Yu Lee, and Andrew Rabinovich.
\newblock Gradnorm: Gradient normalization for adaptive loss balancing in deep
  multitask networks.
\newblock In {\em International Conference on Machine Learning}, pages
  794--803. PMLR, 2018.

\bibitem{chen2022obj2seq}
Zhiyang Chen, Yousong Zhu, Zhaowen Li, Fan Yang, Wei Li, Haixin Wang, Chaoyang
  Zhao, Liwei Wu, Rui Zhao, Jinqiao Wang, et~al.
\newblock Obj2seq: Formatting objects as sequences with class prompt for visual
  tasks.
\newblock {\em arXiv preprint arXiv:2209.13948}, 2022.

\bibitem{cheng2021per}
Bowen Cheng, Alex Schwing, and Alexander Kirillov.
\newblock Per-pixel classification is not all you need for semantic
  segmentation.
\newblock {\em Advances in Neural Information Processing Systems},
  34:17864--17875, 2021.

\bibitem{clark2019bam}
Kevin Clark, Minh-Thang Luong, Urvashi Khandelwal, Christopher~D Manning, and
  Quoc~V Le.
\newblock Bam! born-again multi-task networks for natural language
  understanding.
\newblock {\em arXiv preprint arXiv:1907.04829}, 2019.

\bibitem{Crawshaw2020MultiTaskLW}
Michael Crawshaw.
\newblock Multi-task learning with deep neural networks: A survey.
\newblock {\em arXiv preprint arXiv:2009.09796}, 2020.

\bibitem{desideri2012multiple}
Jean-Antoine D{\'e}sid{\'e}ri.
\newblock Multiple-gradient descent algorithm (mgda) for multiobjective
  optimization.
\newblock {\em Comptes Rendus Mathematique}, 350(5-6):313--318, 2012.

\bibitem{dosovitskiy2020image}
Alexey Dosovitskiy, Lucas Beyer, Alexander Kolesnikov, Dirk Weissenborn,
  Xiaohua Zhai, Thomas Unterthiner, Mostafa Dehghani, Matthias Minderer, Georg
  Heigold, Sylvain Gelly, et~al.
\newblock An image is worth 16x16 words: Transformers for image recognition at
  scale.
\newblock {\em arXiv preprint arXiv:2010.11929}, 2020.

\bibitem{Eigen2015Predicting}
David Eigen and Rob Fergus.
\newblock Predicting depth, surface normals and semantic labels with a common
  multi-scale convolutional architecture.
\newblock In {\em Proceedings of the IEEE International Conference on Computer
  Vision (ICCV)}, pages 2650--2658, 2015.

\bibitem{Fifty2021EfficientlyIT}
Christopher Fifty, Ehsan Amid, Zhe Zhao, Tianhe Yu, Rohan Anil, and Chelsea
  Finn.
\newblock Efficiently identifying task groupings for multi-task learning.
\newblock {\em Advances in Neural Information Processing Systems (NeurIPS)},
  34, 2021.

\bibitem{Gao2021CLIPAdapterBV}
Peng Gao, Shijie Geng, Renrui Zhang, Teli Ma, Rongyao Fang, Yongfeng Zhang,
  Hongsheng Li, and Yu Qiao.
\newblock Clip-adapter: Better vision-language models with feature adapters.
\newblock {\em arXiv preprint arXiv:2110.04544}, 2021.

\bibitem{gao2019nddr}
Yuan Gao, Jiayi Ma, Mingbo Zhao, Wei Liu, and Alan~L Yuille.
\newblock Nddr-cnn: Layerwise feature fusing in multi-task cnns by neural
  discriminative dimensionality reduction.
\newblock In {\em Proceedings of the IEEE/CVF Conference on Computer Vision and
  Pattern Recognition}, pages 3205--3214, 2019.

\bibitem{Ghiasi2021Multi}
Golnaz Ghiasi, Barret Zoph, Ekin~D. Cubuk, Quoc~V. Le, and Tsung-Yi Lin.
\newblock Multi-task self-training for learning general representations.
\newblock In {\em Proceedings of the IEEE/CVF International Conference on
  Computer Vision (ICCV)}, pages 8856--8865, 2021.

\bibitem{He2017Mask}
Kaiming He, Georgia Gkioxari, Piotr Dollar, and Ross Girshick.
\newblock Mask r-cnn.
\newblock In {\em Proceedings of the IEEE International Conference on Computer
  Vision (ICCV)}, pages 2961--2969, 2017.

\bibitem{he2016deep}
Kaiming He, Xiangyu Zhang, Shaoqing Ren, and Jian Sun.
\newblock Deep residual learning for image recognition.
\newblock In {\em Proceedings of the IEEE conference on computer vision and
  pattern recognition}, pages 770--778, 2016.

\bibitem{he2022hyperprompt}
Yun He, Steven Zheng, Yi Tay, Jai Gupta, Yu Du, Vamsi Aribandi, Zhe Zhao,
  YaGuang Li, Zhao Chen, Donald Metzler, et~al.
\newblock Hyperprompt: Prompt-based task-conditioning of transformers.
\newblock In {\em International Conference on Machine Learning}, pages
  8678--8690. PMLR, 2022.

\bibitem{hou2019learning}
Yuenan Hou, Zheng Ma, Chunxiao Liu, and Chen~Change Loy.
\newblock Learning lightweight lane detection cnns by self attention
  distillation.
\newblock In {\em Proceedings of the IEEE/CVF international conference on
  computer vision}, pages 1013--1021, 2019.

\bibitem{ishihara2021multi}
Keishi Ishihara, Anssi Kanervisto, Jun Miura, and Ville Hautamaki.
\newblock Multi-task learning with attention for end-to-end autonomous driving.
\newblock In {\em Proceedings of the IEEE/CVF Conference on Computer Vision and
  Pattern Recognition}, pages 2902--2911, 2021.

\bibitem{jia2022visual}
Menglin Jia, Luming Tang, Bor-Chun Chen, Claire Cardie, Serge Belongie, Bharath
  Hariharan, and Ser-Nam Lim.
\newblock Visual prompt tuning.
\newblock {\em arXiv preprint arXiv:2203.12119}, 2022.

\bibitem{Kendall2018Multi}
Alex Kendall, Yarin Gal, and Roberto Cipolla.
\newblock Multi-task learning using uncertainty to weigh losses for scene
  geometry and semantics.
\newblock In {\em Proceedings of the IEEE/CVF Conference on Computer Vision and
  Pattern Recognition (CVPR)}, pages 7482--7491, 2018.

\bibitem{kirillov2019panoptic}
Alexander Kirillov, Ross Girshick, Kaiming He, and Piotr Doll{\'a}r.
\newblock Panoptic feature pyramid networks.
\newblock In {\em Proceedings of the IEEE/CVF Conference on Computer Vision and
  Pattern Recognition (CVPR)}, pages 6399--6408, 2019.

\bibitem{kokkinos2017ubernet}
Iasonas Kokkinos.
\newblock Ubernet: Training a universal convolutional neural network for low-,
  mid-, and high-level vision using diverse datasets and limited memory.
\newblock In {\em Proceedings of the IEEE conference on computer vision and
  pattern recognition}, pages 6129--6138, 2017.

\bibitem{li2021prefix}
Xiang~Lisa Li and Percy Liang.
\newblock Prefix-tuning: Optimizing continuous prompts for generation.
\newblock {\em arXiv preprint arXiv:2101.00190}, 2021.

\bibitem{liang2022effective}
Xiwen Liang, Yangxin Wu, Jianhua Han, Hang Xu, Chunjing Xu, and Xiaodan Liang.
\newblock Effective adaptation in multi-task co-training for unified autonomous
  driving.
\newblock {\em arXiv preprint arXiv:2209.08953}, 2022.

\bibitem{Likhosherstov2021PolyViTCV}
Valerii Likhosherstov, Anurag Arnab, Krzysztof Choromanski, Mario Lucic, Yi
  Tay, Adrian Weller, and Mostafa Dehghani.
\newblock Polyvit: Co-training vision transformers on images, videos and audio.
\newblock {\em arXiv preprint arXiv:2111.12993}, 2021.

\bibitem{likhosherstov2021polyvit}
Valerii Likhosherstov, Anurag Arnab, Krzysztof Choromanski, Mario Lucic, Yi
  Tay, Adrian Weller, and Mostafa Dehghani.
\newblock Polyvit: Co-training vision transformers on images, videos and audio.
\newblock {\em arXiv preprint arXiv:2111.12993}, 2021.

\bibitem{lin2017feature}
Tsung-Yi Lin, Piotr Doll{\'a}r, Ross Girshick, Kaiming He, Bharath Hariharan,
  and Serge Belongie.
\newblock Feature pyramid networks for object detection.
\newblock In {\em Proceedings of the IEEE conference on computer vision and
  pattern recognition}, pages 2117--2125, 2017.

\bibitem{lin2019pareto}
Xi Lin, Hui-Ling Zhen, Zhenhua Li, Qing-Fu Zhang, and Sam Kwong.
\newblock Pareto multi-task learning.
\newblock {\em Advances in neural information processing systems}, 32, 2019.

\bibitem{liu2021pre}
Pengfei Liu, Weizhe Yuan, Jinlan Fu, Zhengbao Jiang, Hiroaki Hayashi, and
  Graham Neubig.
\newblock Pre-train, prompt, and predict: A systematic survey of prompting
  methods in natural language processing.
\newblock {\em arXiv preprint arXiv:2107.13586}, 2021.

\bibitem{liu2019end}
Shikun Liu, Edward Johns, and Andrew~J Davison.
\newblock End-to-end multi-task learning with attention.
\newblock In {\em Proceedings of the IEEE/CVF conference on computer vision and
  pattern recognition}, pages 1871--1880, 2019.

\bibitem{liu2021gpt}
Xiao Liu, Yanan Zheng, Zhengxiao Du, Ming Ding, Yujie Qian, Zhilin Yang, and
  Jie Tang.
\newblock Gpt understands, too.
\newblock {\em arXiv preprint arXiv:2103.10385}, 2021.

\bibitem{liu2021swin}
Ze Liu, Yutong Lin, Yue Cao, Han Hu, Yixuan Wei, Zheng Zhang, Stephen Lin, and
  Baining Guo.
\newblock Swin transformer: Hierarchical vision transformer using shifted
  windows.
\newblock In {\em Proceedings of the IEEE/CVF International Conference on
  Computer Vision}, pages 10012--10022, 2021.

\bibitem{marfoq2021federated}
Othmane Marfoq, Giovanni Neglia, Aur{\'e}lien Bellet, Laetitia Kameni, and
  Richard Vidal.
\newblock Federated multi-task learning under a mixture of distributions.
\newblock {\em Advances in Neural Information Processing Systems}, 34, 2021.

\bibitem{misra2016cross}
Ishan Misra, Abhinav Shrivastava, Abhinav Gupta, and Martial Hebert.
\newblock Cross-stitch networks for multi-task learning.
\newblock In {\em Proceedings of the IEEE conference on computer vision and
  pattern recognition}, pages 3994--4003, 2016.

\bibitem{nie2022pro}
Xing Nie, Bolin Ni, Jianlong Chang, Gaomeng Meng, Chunlei Huo, Zhaoxiang Zhang,
  Shiming Xiang, Qi Tian, and Chunhong Pan.
\newblock Pro-tuning: Unified prompt tuning for vision tasks.
\newblock {\em arXiv preprint arXiv:2207.14381}, 2022.

\bibitem{Radford2021LearningTV}
Alec Radford, Jong~Wook Kim, Chris Hallacy, Aditya Ramesh, Gabriel Goh,
  Sandhini Agarwal, Girish Sastry, Amanda Askell, Pamela Mishkin, Jack Clark,
  Gretchen Krueger, and Ilya Sutskever.
\newblock Learning transferable visual models from natural language
  supervision.
\newblock In {\em International Conference on Machine Learning (ICML)}, pages
  8748--8763, 2021.

\bibitem{rao2022denseclip}
Yongming Rao, Wenliang Zhao, Guangyi Chen, Yansong Tang, Zheng Zhu, Guan Huang,
  Jie Zhou, and Jiwen Lu.
\newblock Denseclip: Language-guided dense prediction with context-aware
  prompting.
\newblock In {\em Proceedings of the IEEE/CVF Conference on Computer Vision and
  Pattern Recognition}, pages 18082--18091, 2022.

\bibitem{Ren2015Faster}
Shaoqing Ren, Kaiming He, Ross Girshick, and Jian Sun.
\newblock Faster r-cnn: Towards real-time object detection with region proposal
  networks.
\newblock {\em Advances in Neural Information Processing Systems (NeurIPS)},
  28, 2015.

\bibitem{ruder2019latent}
Sebastian Ruder, Joachim Bingel, Isabelle Augenstein, and Anders S{\o}gaard.
\newblock Latent multi-task architecture learning.
\newblock In {\em Proceedings of the AAAI Conference on Artificial
  Intelligence}, volume~33, pages 4822--4829, 2019.

\bibitem{sener2018multi}
Ozan Sener and Vladlen Koltun.
\newblock Multi-task learning as multi-objective optimization.
\newblock {\em Advances in neural information processing systems}, 31, 2018.

\bibitem{Standley2020WhichTS}
Trevor~Scott Standley, Amir~Roshan Zamir, Dawn Chen, Leonidas~J. Guibas,
  Jitendra Malik, and Silvio Savarese.
\newblock Which tasks should be learned together in multi-task learning?
\newblock In {\em International Conference on Machine Learning (ICML)}, pages
  9120--9132, 2020.

\bibitem{Sun2021Sparse}
Peize Sun, Rufeng Zhang, Yi Jiang, Tao Kong, Chenfeng Xu, Wei Zhan, Masayoshi
  Tomizuka, Lei Li, Zehuan Yuan, Changhu Wang, and Ping Luo.
\newblock Sparse r-cnn: End-to-end object detection with learnable proposals.
\newblock In {\em Proceedings of the IEEE/CVF Conference on Computer Vision and
  Pattern Recognition (CVPR)}, pages 14454--14463, 2021.

\bibitem{teichmann2016multinet}
Marvin Teichmann, Michael Weber, Marius Zoellner, Roberto Cipolla, and Raquel
  Urtasun.
\newblock Multinet: Real-time joint semantic reasoning for autonomous driving.
\newblock In {\em IEEE Intelligent Vehicles Symposium (IV)}, pages 1013--1020,
  2018.

\bibitem{vandenhende2020mti}
Simon Vandenhende, Stamatios Georgoulis, and Luc~Van Gool.
\newblock Mti-net: Multi-scale task interaction networks for multi-task
  learning.
\newblock In {\em European Conference on Computer Vision}, pages 527--543.
  Springer, 2020.

\bibitem{vandenhende2021multi}
Simon Vandenhende, Stamatios Georgoulis, Wouter Van~Gansbeke, Marc Proesmans,
  Dengxin Dai, and Luc Van~Gool.
\newblock Multi-task learning for dense prediction tasks: A survey.
\newblock {\em IEEE transactions on pattern analysis and machine intelligence},
  2021.

\bibitem{wang2018glue}
Alex Wang, Amanpreet Singh, Julian Michael, Felix Hill, Omer Levy, and Samuel~R
  Bowman.
\newblock Glue: A multi-task benchmark and analysis platform for natural
  language understanding.
\newblock {\em arXiv preprint arXiv:1804.07461}, 2018.

\bibitem{wang2018lanenet}
Ze Wang, Weiqiang Ren, and Qiang Qiu.
\newblock Lanenet: Real-time lane detection networks for autonomous driving.
\newblock {\em arXiv preprint arXiv:1807.01726}, 2018.

\bibitem{wang2022dualprompt}
Zifeng Wang, Zizhao Zhang, Sayna Ebrahimi, Ruoxi Sun, Han Zhang, Chen-Yu Lee,
  Xiaoqi Ren, Guolong Su, Vincent Perot, Jennifer Dy, et~al.
\newblock Dualprompt: Complementary prompting for rehearsal-free continual
  learning.
\newblock {\em arXiv preprint arXiv:2204.04799}, 2022.

\bibitem{wu2021yolop}
Dong Wu, Manwen Liao, Weitian Zhang, and Xinggang Wang.
\newblock Yolop: You only look once for panoptic driving perception.
\newblock {\em arXiv preprint arXiv:2108.11250}, 2021.

\bibitem{Xiao2018UnifiedPP}
Tete Xiao, Yingcheng Liu, Bolei Zhou, Yuning Jiang, and Jian Sun.
\newblock Unified perceptual parsing for scene understanding.
\newblock In {\em European Conference on Computer Vision (ECCV)}, pages
  418--434, 2018.

\bibitem{xiao2018unified}
Tete Xiao, Yingcheng Liu, Bolei Zhou, Yuning Jiang, and Jian Sun.
\newblock Unified perceptual parsing for scene understanding.
\newblock In {\em Proceedings of the European Conference on Computer Vision
  (ECCV)}, pages 418--434, 2018.

\bibitem{xie2021cotr}
Yutong Xie, Jianpeng Zhang, Chunhua Shen, and Yong Xia.
\newblock Cotr: Efficiently bridging cnn and transformer for 3d medical image
  segmentation.
\newblock In {\em International conference on medical image computing and
  computer-assisted intervention}, pages 171--180, 2021.

\bibitem{xu2018pad}
Dan Xu, Wanli Ouyang, Xiaogang Wang, and Nicu Sebe.
\newblock Pad-net: Multi-tasks guided prediction-and-distillation network for
  simultaneous depth estimation and scene parsing.
\newblock In {\em Proceedings of the IEEE Conference on Computer Vision and
  Pattern Recognition}, pages 675--684, 2018.

\bibitem{xu2022multi}
Yangyang Xu, Xiangtai Li, Haobo Yuan, Yibo Yang, Jing Zhang, Yunhai Tong, Lefei
  Zhang, and Dacheng Tao.
\newblock Multi-task learning with multi-query transformer for dense
  prediction.
\newblock {\em arXiv preprint arXiv:2205.14354}, 2022.

\bibitem{Yang2018EndtoendMM}
Zhengyuan Yang, Yixuan Zhang, Jerry Yu, Junjie Cai, and Jiebo Luo.
\newblock End-to-end multi-modal multi-task vehicle control for self-driving
  cars with visual perceptions.
\newblock {\em 24th International Conference on Pattern Recognition (ICPR)},
  pages 2289--2294, 2018.

\bibitem{yao2021cpt}
Yuan Yao, Ao Zhang, Zhengyan Zhang, Zhiyuan Liu, Tat-Seng Chua, and Maosong
  Sun.
\newblock Cpt: Colorful prompt tuning for pre-trained vision-language models.
\newblock {\em arXiv preprint arXiv:2109.11797}, 2021.

\bibitem{yu2020bdd100k}
Fisher Yu, Haofeng Chen, Xin Wang, Wenqi Xian, Yingying Chen, Fangchen Liu,
  Vashisht Madhavan, and Trevor Darrell.
\newblock Bdd100k: A diverse driving dataset for heterogeneous multitask
  learning.
\newblock In {\em Proceedings of the IEEE/CVF conference on computer vision and
  pattern recognition}, pages 2636--2645, 2020.

\bibitem{zhang2022dino}
Hao Zhang, Feng Li, Shilong Liu, Lei Zhang, Hang Su, Jun Zhu, Lionel~M Ni, and
  Heung-Yeung Shum.
\newblock Dino: Detr with improved denoising anchor boxes for end-to-end object
  detection.
\newblock {\em arXiv preprint arXiv:2203.03605}, 2022.

\bibitem{zhang2018joint}
Zhenyu Zhang, Zhen Cui, Chunyan Xu, Zequn Jie, Xiang Li, and Jian Yang.
\newblock Joint task-recursive learning for semantic segmentation and depth
  estimation.
\newblock In {\em Proceedings of the European Conference on Computer Vision
  (ECCV)}, pages 235--251, 2018.

\bibitem{zhong2021factual}
Zexuan Zhong, Dan Friedman, and Danqi Chen.
\newblock Factual probing is [mask]: Learning vs. learning to recall.
\newblock {\em arXiv preprint arXiv:2104.05240}, 2021.

\bibitem{Zhou2021LearningTP}
Kaiyang Zhou, Jingkang Yang, Chen~Change Loy, and Ziwei Liu.
\newblock Learning to prompt for vision-language models.
\newblock {\em arXiv preprint arXiv:2109.01134}, 2021.

\bibitem{zhou2022conditional}
Kaiyang Zhou, Jingkang Yang, Chen~Change Loy, and Ziwei Liu.
\newblock Conditional prompt learning for vision-language models.
\newblock In {\em Proceedings of the IEEE/CVF Conference on Computer Vision and
  Pattern Recognition}, pages 16816--16825, 2022.

\end{thebibliography}
}

\newpage
\appendix

\section{Experimental Setup}
\subsection{Implementation Details}
Here we further provide detailed experimental settings in our paper.
Class numbers for object detection, semantic segmentation, drivable area segmentation, and lane detection are 9, 19, 2, and 1 respectively. We remove the train class as in \cite{wu2021yolop} for object detection. For lane detection, we follow \cite{hou2019learning} to preprocess lane line annotations. Loss weights for object detection, semantic segmentation, drivable area segmentation, and lane detection are fixed as 1, 2, 2, and 2 respectively.
All experiments are conducted on servers with 8 Nvidia V100 GPUs and Intel Xeon Platinum 8168 CPU (2.70GHz).

\subsection{More Details on Dataset}
BDD100k dataset \cite{yu2020bdd100k} contains multiple tasks. Here we focus on object detection (OD), semantic segmentation (SS), drivable area segmentation (DA), and lane detection (LD). In BDD100K, 70k training images are labeled for object detection, drivable area segmentation, and lane detection, and only 7k training images are labeled for semantic segmentation.

\section{More Investigations}
In this section, we present more analyses of popular multi-task learning methods.

\subsection{Task Scheduling}
Here we analyze task scheduling methods on disjoint-balance settings, whose results are shown in Table 1 in the main paper. Note that the full setting contains almost complete annotations except semantic segmentation, thus it is not suitable for task scheduling. Since the data of all tasks in the disjoint-balance setting is balanced and non-overlapped, Uniform sampler \cite{Likhosherstov2021PolyViTCV} and Weighted sampler \cite{Likhosherstov2021PolyViTCV} are equivalent.
As shown in Table 1 in the main paper, task sampling methods (i.e., Uniform sampler \cite{Likhosherstov2021PolyViTCV} and Round-robin \cite{Likhosherstov2021PolyViTCV}) perform better than Zeroing loss \cite{Xiao2018UnifiedPP} by a large margin on segmentation-based tasks, but get worse in object detection. We hypothesize that negative transfer still exists among these approaches, and training one task per step may lead to forgetting to some extent.

\subsection{Partial-label Learning}
As shown in Table 1 in the main paper, pseudo labeling \cite{Ghiasi2021Multi} surpasses Zeroing loss \cite{Xiao2018UnifiedPP} on almost all tasks, especially on semantic segmentation. However, the improvement in drivable area segmentation and lane detection under the full setting is not obvious, since there are less unlabeled data on these tasks.

\begin{table*}[t]
    \centering
    \caption{Compare with LV-Adapter under the disjoint-balance setting.}
    \begin{tabular}{c|ccc|c|c|c|c|c}
        \toprule
        Method & mAP & AP50 & AP75 & mIoU (SS) & mIoU (DA) & IoU (LD) & GFLOPs & Params \\
        \midrule
        LV-Adapter \cite{liang2022effective} & 24.6 & 47.4 & 21.9 & \textbf{61.8} & 80.6 & - & 415 & 200M \\
        VE-Prompt (Ours) & \textbf{26.8} & \textbf{51.2} & \textbf{23.8} & 58.3 & \textbf{86.8} & \textbf{22.1} & 401 & 60M \\
        \bottomrule
    \end{tabular}
    \label{tab:lvadapter}
\end{table*}

\subsection{Task Balancing}
We choose pseudo labeling as the baseline since task-balancing methods are more suitable in settings with complete labels. Fixed denotes fixed loss weights for all tasks during training. As shown in Table 2 in the main paper, Uncertainty \cite{Kendall2018Multi} performs better than Fixed \cite{Ghiasi2021Multi} under the full settings overall, while the performance of GradNorm \cite{chen2018gradnorm} degrades significantly.
Interestingly, Fixed performs slightly better than Uncertainty under the disjoint-balance setting, which indicates that Uncertainty is not suitable for all data split settings. GradNorm and MGDA \cite{sener2018multi} perform poorly overall, showing that these task-balancing methods are not suitable for autonomous driving.
Especially, GradNorm uses the last shared layer of weights to compute gradient norm in its paper, thus we adopt the last layer of $P_5$ in the neck. We also choose the whole shared encoder to implement GradNorm, which is denoted as GradNorm$^*$, improving the original one by a large margin under the disjoint-balance setting ($+$10.8 in Avg.). This indicates that the selection of network weights for computing GradNorm is important.
Interestingly, MGDA consistently achieves the best result on lane detection, indicating that it suffers from the heavy negative transfer. Since it takes a long time to train MGDA, we did not implement it under the full setting for the time limit.

\begin{table*}[t]
    \centering
    \caption{Comparison of fixed and trainable task-specific prompts under the disjoint-balance setting.}
    \begin{tabular}{c|ccc|c|c|c}
        \toprule
        Fixed & mAP & AP50 & AP75 & mIoU (SS) & mIoU (DA) & IoU (LD) \\
        \midrule
        \cmark & 33.3 & 55.3 & 32.5 & 61.1 & 87.2 & 22.1 \\
        \xmark & \textbf{33.9} & \textbf{56.6} & \textbf{33.7} & \textbf{61.2} & \textbf{87.4} & \textbf{22.2} \\
        \bottomrule
    \end{tabular}
    \label{tab:fix_prompt}
\end{table*}

\begin{figure}[t]
    \centering
    \includegraphics[width=1.0\linewidth]{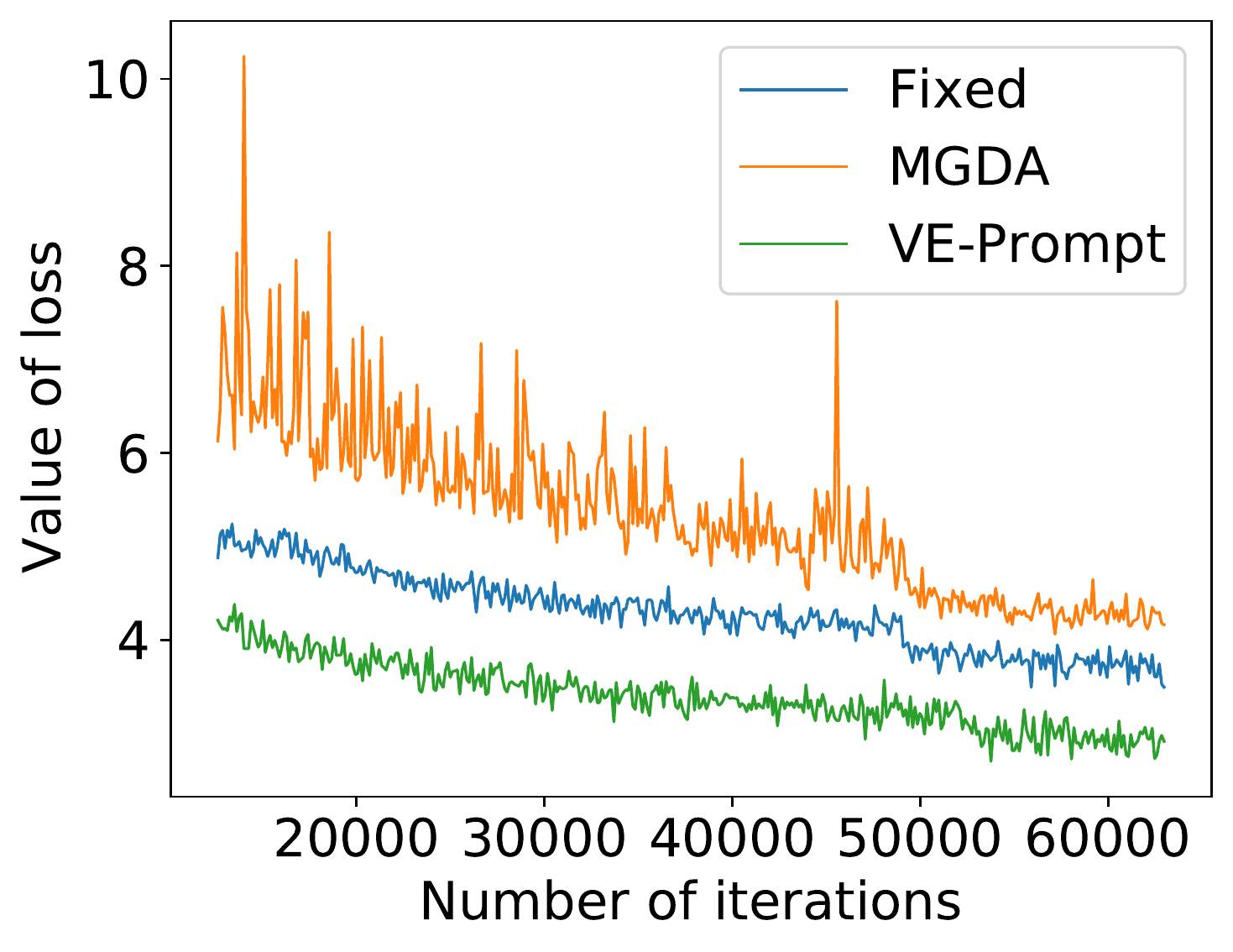}
    \caption{We provide the loss changes of all tasks during training under the disjoint-balance setting. From the curve, we find out that our VE-Prompt can achieve faster and better convergence.}
    \label{fig:loss_ann}
\end{figure}

\begin{figure*}[t]
    \centering
    \includegraphics[width=1.0\linewidth]{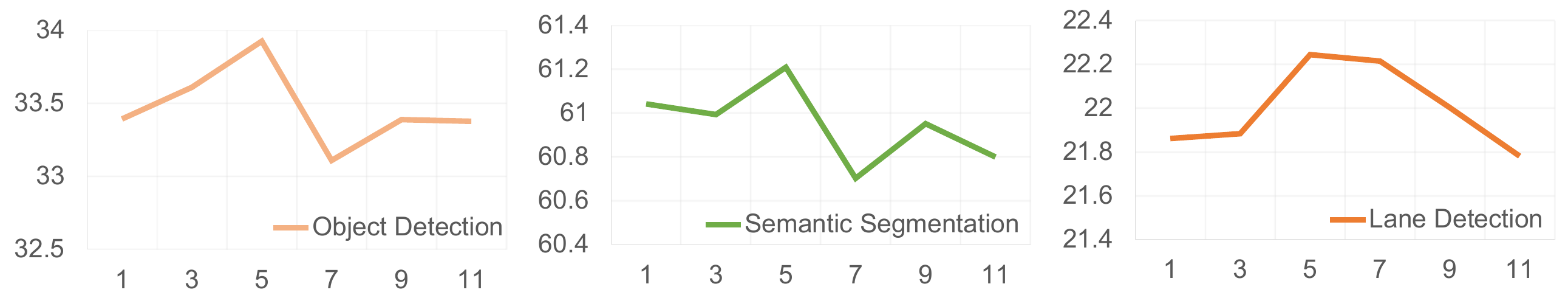}
    \caption{Ablation study of different numbers of visual exemplars under the disjoint-balance setting. The x-axis represents the number of visual exemplars, and the y-axis indicates mAP or mIoU.}
    \label{fig:num_prompt}
\end{figure*}

In summary, most existing multi-task learning methods suffer from poor performances under the real-world scenarios of autonomous driving since they are not designed to handle unified perception in self-driving. Therefore, it is extremely important and urgent to develop applicable multi-task methods for autonomous driving.

\section{Compare with LV-Adapter}
We conduct experiments to compare with recent LV-Adapter \cite{liang2022effective}, which tackles three tasks, as in Table \ref{tab:lvadapter}. The class number of object detection is 10 in LV-Adapter, and the backbone is Res50 \cite{he2016deep}. Here we use the same setting as in LV-Adapter, and present results under the disjoint-balance setting in Table \ref{tab:lvadapter}. Note that data splits of object detection, semantic segmentation, and drivable area segmentation are the same as in LV-Adapter. Results show that our proposed VE-Prompt performs better than LV-Adapter on object detection and semantic segmentation by a large margin ($+$2.2 in mAP and $+$6.2 in mIoU (DA)). Meanwhile, our method gets competitive results in lane detection compared with the Swin-Tiny \cite{liu2021swin} backbone as in Table 3 in the main paper. LV-Adapter adopts MaskFormer \cite{cheng2021per}, which is a stronger baseline, to generate pseudo labels for semantic segmentation, while VE-Prompt chooses Semantic FPN \cite{kirillov2019panoptic} as the teacher model. Therefore, the improvement of semantic segmentation for LV-Adapter may come from high-quality pseudo labels.
The number of parameters in the proposed VE-Prompt is much less than that of LV-Adapter as in Table \ref{tab:lvadapter}. We also report GFLOPs on the same V100 NVIDIA GPU for a fair comparison. Results show that our method is more efficient and effective overall.

\section{More Ablation Studies}
\subsection{Influence of Fixed Prompts}
The task-specific prompts are not fixed during training in VE-Prompt. We also conduct experiments to verify the effectiveness of trainable task-specific prompts as in Table \ref{tab:fix_prompt}. Results show that the model with trainable task-specific prompts performs better on all four tasks.

\subsection{Number of Exemplars}
Here we compare different configurations of the number of visual exemplars. The number of visual exemplars for different tasks is $n_1$, $n_2$, $n_3$, and $n_4$. We keep them equal for simplification. As shown in Figure \ref{fig:num_prompt}, the model performs better when $n_1 = n_2 = n_3 = n_5 = 5$, thus we set the number of visual exemplars as 5 in our final model.

\subsection{Loss Analysis}
We also analyze the loss changes of VE-Prompt and the baseline under the disjoint-balance setting as in Figure \ref{fig:loss_ann}. From the loss curves, we conclude that our VE-Prompt achieves consistent faster and better convergence during training. Note that loss weights for all tasks in Fixed and VE-Prompt here are set as 1 for a fair comparison.

\subsection{Comparisons with Alternative Options}
We present the results of VE-Prompt with some alternative options as in Table \ref{tab:option}. Results show that VE-Prompt with Uncertainty can improve Uncertainty on all tasks, and VE-Prompt with Fixed performs better than VE-Prompt with Uncertainty.

\begin{table}[h]
    \centering
    \caption{Results of VE-Prompt with alternative options under the disjoint-balance setting.}
    \resizebox{1.0\linewidth}{!}{
    \begin{tabular}{c|c|c|c|c|c}
        \toprule
        Model & mAP & mIoU (SS) & mIoU (DA) & IoU (LD) & $\Delta_{MTL}(\%)$ \\
        \midrule
        Uncertainty & 31.2 & 59.9 & 87.0 & 22.2 & +1.66 \\
        VE-Prompt (Uncertainty) & \textbf{32.9} & \textbf{60.6} & \textbf{87.9} & \textbf{22.5} & \textbf{+4.04} \\
        \midrule
        Fixed & 31.3 & 60.2 & 87.0 & \textbf{22.2} & +1.87 \\
        VE-Prompt (Fixed) & \textbf{33.9} & \textbf{61.2} & \textbf{87.4} & \textbf{22.2} & \textbf{+4.72} \\
        \bottomrule
    \end{tabular}}
    \label{tab:option}
\end{table}

\section{Experiments on NuImages Dataset}
We also conduct experiments on nuImages dataset\footnote{\url{https://www.nuscenes.org/nuimages}}, which covers two tasks, object detection and semantic segmentation. Results are shown in Table \ref{tab:nuimages}, indicating that VE-Prompt performs much better than baselines.

\begin{table}[h]
    \centering
    \caption{Comparisons with multi-task baselines on nuImages.}
    \resizebox{1.0\linewidth}{!}{
    \begin{tabular}{c|ccc|c|c}
        \toprule
        Model & mAP & AP50 & AP75 & mIoU & Avg. \\ %
        \midrule
        Sparse R-CNN based & 50.4 & 76.8 & 54.5 & 53.8 & 52.1 \\
        DINO based & 55.5 & 81.6 & 60.6 & 56.7 & 56.1 \\
        VE-Prompt (Ours) & \textbf{55.8} & \textbf{81.9} & \textbf{60.7} & \textbf{59.1} & \textbf{57.5} \\
        \bottomrule
    \end{tabular}}
    \label{tab:nuimages}
\end{table}

\end{document}